\documentclass[11pt,a4paper]{article}
\usepackage[hyperref]{eacl2021}
\usepackage{times}
\usepackage{latexsym}

\usepackage{graphicx}
\usepackage[ruled, linesnumbered]{algorithm2e}
\usepackage{amssymb,mathtools}
\usepackage{xcolor}
\usepackage{comment}
\usepackage{mathtools}
\usepackage[thinlines]{easytable}
\usepackage{babel,blindtext}

\usepackage{diagbox}
\usepackage{multirow}
\usepackage{longtable}
\usepackage{dblfloatfix}
\usepackage{scrextend}
\usepackage{float}
\usepackage{caption}
\usepackage{url}
\usepackage{multicol}
\usepackage{amsfonts}
\usepackage{subfigure}

\usepackage{array}
\newcolumntype{L}[1]{>{\raggedright\let\newline\\\arraybackslash\hspace{0pt}}m{#1}}
\newcolumntype{C}[1]{>{\centering\let\newline\\\arraybackslash\hspace{0pt}}m{#1}}
\newcolumntype{R}[1]{>{\raggedleft\let\newline\\\arraybackslash\hspace{0pt}}m{#1}}

\usepackage{microtype}

\aclfinalcopy 

\title{NLP-CUET@DravidianLangTech-EACL2021: Investigating Visual and Textual Features to Identify Trolls from Multimodal Social Media Memes}

\author{Eftekhar Hossain{*}, Omar Sharif{\textdagger}  and Mohammed Moshiul Hoque{\textdagger}\\
    {\textdagger}Department of Computer Science and Engineering \\
    {*}Department of Electronics and Telecommunication Engineering\\
    Chittagong University of Engineering and Technology, Bangladesh \\
  \texttt{\{eftekhar.hossain, omar.sharif, moshiul\_240\}@cuet.ac.bd}\\

  }

\date{}

\begin{document}
\maketitle
\begin{abstract}
In the past few years, the meme has become a new way of communication on the Internet. As memes are the images with embedded text, it can quickly spread hate, offence and violence. Classifying memes are very challenging because of their multimodal nature and region-specific interpretation. A shared task is organized to develop models that can identify trolls from multimodal social media memes. This work presents a computational model that we have developed as part of our participation in the task. Training data comes in two forms: an image with embedded Tamil code-mixed text and an associated caption given in English. We investigated the visual and textual features using CNN, VGG16, Inception, Multilingual-BERT, XLM-Roberta, XLNet models. Multimodal features are extracted by combining image (CNN, ResNet50, Inception) and text (Long short term memory network) features via early fusion approach. Results indicate that the textual approach with XLNet achieved the highest weighted $f_1$-score of $0.58$, which enabled our model to secure $3^{rd}$ rank in this task. 
\end{abstract}

\section{Introduction}
With the Internet's phenomenal growth,  social media has become a platform for sharing information, opinion, feeling, expressions, and ideas. Most users enjoy the liberty to post or share contents in such virtual platforms without any legal authority intervention or moderation. Some people misuse this freedom and take social platforms to spread negativity, threat and offence against individuals or communities in various ways. One such way is making and sharing troll memes to provoke, offend and demean a group or race on the Internet \citep{mojica-de-la-vega-ng-2018-modeling}. Although memes meant to be sarcastic or humorous, sometimes it becomes aggressive, threatening and abusive \citep{suryawanshi-etal-2020-multimodal}. Till to date, extensive research has been conducted to detect hate, hostility, and aggression from a single modality such as image or text \citep{trac-2020-trolling}. Identification of troll, offence, abuse by analyzing the combined information of visual and textual modalities is still an unexplored research avenue in natural language processing (NLP). Classifying memes from multi-model data is a challenging task since memes express sarcasm and humour implicitly. One meme may not be a troll if we consider only image or text associated with it. However, it can be a troll if it considers both text and image modalities. Such implicit meaning of memes, use of sarcastic, ambiguous and humorous terms and absence of baseline algorithms that take care of multiple modalities are the primary concerns of categorizing multimodal memes. Features from multiple modalities (i.e image, text) have been exploited in many works to solve these problems \citep{suryawanshi-etal-2020-multimodal, pranesh2020memesem}. We plan to address this issue by using transfer learning as these models have better generalization capability than the models trained on small dataset. This work is a little effort to compensate for the existing deficiency of assign task with the following contributions:
\begin{itemize}
    \item Developed a classifier model using XLNet to identify trolls from multimodal social media memes.
    \item Investigate the performance of various transfer learning techniques with the benchmark experimental evaluation by exploring visual, textual and multimodal features of the data.
\end{itemize}

\section{Related Work}
In the past few years, trolling, aggression, hostile and abusive language detection from social media data has been studied widely by NLP experts \cite{trac-2020-trolling, sharif2021combating, 10.1145/3368567.3368584}. Majority of these researches carried out concerning the textual information alone \citep{alw-2020-online}. However, a meme existence can be found in a basic image, text embedded in image or image with sarcastic caption. Very few researches have investigated both textual and visual features to classify meme, troll, offence and aggression. \citet{suryawanshi-etal-2020-multimodal} developed a multimodal dataset to identify offensive memes. Authors have combined text and image both modalities to detect an offensive meme. Their model obtained $f_1$-score of $0.50$ in a multimodal setting with LSTM and VGG16 network. A unimodal framework utilized the image features to detect offensive content in image \citep{gandhi2019image}. A multimodal framework is proposed by \citep{pranesh2020memesem} to analyze the underlying sentiment of memes. They used VGG19 model pre-trained on ImageNet and bidirectional encoder representations from transformers (BERT) language model to capture both textual and visual features. \citet{wang-wen-2015-cheezburger} have combined visual and textual information to predict and generate a suitable description for memes. \citet{Du_Masood_Joseph_2020} introduced a model to identify the image with text (IWT) memes that spread hate, offence or misinformation. \citep{suryawanshi-etal-2020-dataset} built a dataset `TamilMemes' to detect troll from memes. In `TamilMemes' text is embedded in the image, each image is labelled as either `troll' or `not-troll'. To perform classification, the authors utilized image-based features, but their method did not achieve credible performance. 

\section{Task and Dataset Descriptions}
Meme as the troll is an image that has offensive or sarcastic text embedded into it and intent to demean, provoke or offend an individual or a group \citep{suryawanshi-etal-2020-dataset}. An image itself can also be a troll meme without any embedded text into it. In this task, we aim to detect troll meme from an image and its associated caption. Task organizers\footnote{https://competitions.codalab.org/competitions/27651} provided a dataset of troll memes for Tamil language \citep{dravidiantrollmeme-eacl}. The dataset contains two parts: an image with embedded Tamil code-mixed text, and a caption. Each instance of the dataset labelled as either `troll' and `not-troll'. Dataset divided into train, validation and test sets. Statistics of the dataset for each class given in table \ref{table1}. Dataset divided into train, validation and test sets. Dataset is imbalanced where several data in the ’troll’ class is much higher than the `not-troll’ class. 

\begin{table}[h!]
\centering
\begin{tabular}{c|cc}
\hline

\hline
\textbf{Data Sets}&\textbf{Troll}&\textbf{Not-Troll}\\
\hline
Train & 1026 & 814 \\
Valid & 256 & 204  \\          
Test & 395  & 272 \\               
\hline
Total & 1677 & 1290\\
\hline
\end{tabular}
\caption{\label{table1}Dataset summary
}
\end{table}

Participants are allowed to use image, caption or both to perform the classification task. We used image, text, and multimodal (i.e., image + text) features to address the assigned task (detail analysis is discussed in Section~\ref{Method}). To understand the properties of the text captions, we further investigated the caption of train sets. Outcomes of the investigation summarized in table \ref{table2}. `troll’ class consists of 7789 unique words with an average of 12.45 words per caption. On the other hand, `not-troll’ class contains 2818 unique words with caption length of 5.40 words on average. Majority of the texts are short and presented sarcastically. As it is challenging to capture the semantics from the short texts, therefore models could not segregate troll ones from the not-troll's effectively. Sample examples from the training set showed in table \ref{table3}. The top image denotes a troll meme, whereas the bottom one represents a typical meme with their associated caption.

\begin{table}[h!]
\centering
\begin{tabular}{C{2.5cm}|cc}
\hline

\hline
\textbf{Attributes}&\textbf{Troll}&\textbf{Not-Troll}\\
\hline
Total words& 12781 & 4402 \\
Unique words & 7789 & 2818  \\         
Max. caption length (in word)& 59  & 29 \\
Avg. words (per caption) & 12.45 & 5.40 \\
\hline            
\end{tabular}
\caption{\label{table2} Training set statistics.
}
\end{table}

\begin{table}[h!]
\centering
\begin{tabular}{C{5cm}L{1.5cm}}
\hline
\textbf{Image}&\textbf{Caption}\\

\includegraphics[height=3.5cm, width=0.3\textwidth]{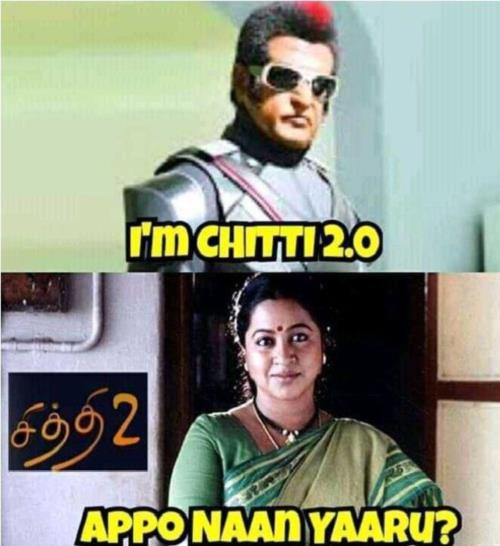}& I;M CHITTI 2.0  APPO NAAN YAARU?E\\
 \includegraphics[height=3.5cm, width=0.3\textwidth]{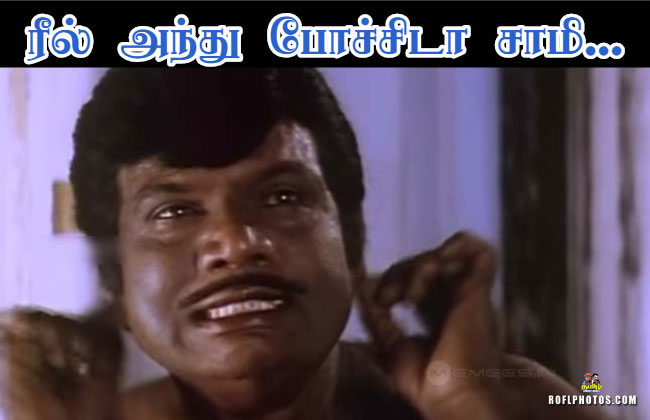} & Ivaingak- ittarundhu yepidi thapikka porenanu theriyalaye\\
\hline
\end{tabular}
\caption{\label{table3} Sample examples of the training set. First one is `troll' meme and second is `not troll' meme.
}
\end{table}

\section{Methodology}
\label{Method}
\begin{figure}[h!]
\centering
    \subfigure[Visual]{\includegraphics[height=1.9cm,width=0.49\linewidth]{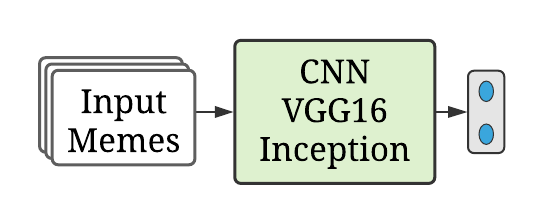}}
    \subfigure[Textual]{\includegraphics[height=1.9cm, width=0.49\linewidth]{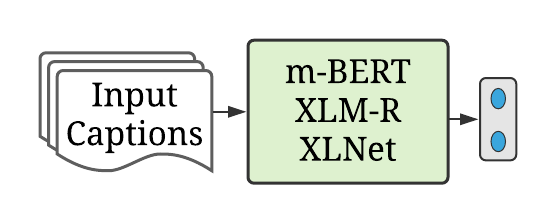}}
    \subfigure[Multimodal]{\includegraphics[ width=0.8\linewidth]{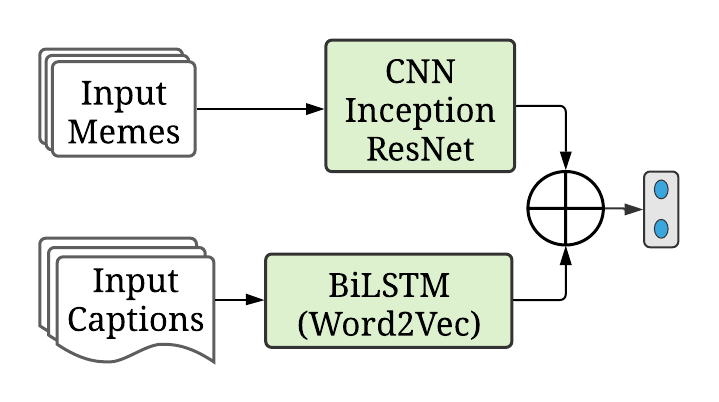}}

 \caption{Abstract process of troll memes classification}
 \label{abstract-view}

\end{figure}

The prime concern of this work is to classify trolling from the multimodal memes. Initially, the investigation begins with accounting only the images' visual features where different CNN architectures will use. The textual features will consider in the next and apply the transformer-based model (i.e. m-BERT, XLM-R, XLNet) for the classification task. Finally, we investigate the effect of combined visual and textual features and compare its performance with the other approaches. Figure \ref{abstract-view} shows the abstract view of the employed techniques.

\subsection{Visual Approach} 
A CNN architecture is used to experiment on visual modality. Besides the pre-trained models, VGG16 and Inception are also employed for the classification task. Before feeding into the model, all the images get resized into a dimension of $150\times 150\times 3$.

\paragraph{CNN:} We design a CNN architecture consists of four convolution layers. The first and second layers contained $32$ and $64$ filters, while $128$ filters are used in third and fourth layers. In all layers, convolution is performed by $3\times3$ kernel and used the Relu non-linearity function. To extract the critical features max pooling is performed with $2\times2$ window after every convolution layer. The flattened output of the final convolution layer is feed to a dense layer consisting of $512$ neurons. To mitigate the chance of overfitting a dropout layer is introduced with a dropout rate of $0.1$. Finally, a sigmoid layer is used for the class prediction.

\paragraph{VGG16:} A CNN architecture \citep{simonyan2015deep} is pre-trained on over $14$ million images of $1000$ classes. It has $16$ weighted layers. The uniqueness of these architecture consists of convolution layers where $3\times3$ filter is used with a stride $1$ and `same' padding. Moreover, every convolution layer is followed by maxpool layer of $2\times2$ filter with stride $2$. This layout is consistent throughout the entire architecture. The final layer contains two fully connected (FC) layers, followed by a softmax layer. To accomplish the task, we froze the top layers of VGG16 and fine-tuned it on our images with adding one global average pooling layer followed by an FC layer of $256$ neurons and a sigmoid layer which is used for the class prediction.   
\paragraph{Inception:} 
An efficient deep neural network initially designed for computer vision tasks \citep{szegedy2016rethinking} which use a combination of all the layers namely, $1\times1$, $3\times3$, $5\times5$ convolutional layer. One unique characteristic of these layers is that it allows the internal layers to pick and choose the appropriate filter size to comprehend certain significant information from the images.
We fine-tuned the InceptionV3 \citep{szegedy2016rethinking} network on our images by freezing the top layers. A global average pooling layer is added along with a fully connected layer of $256$ neurons, followed by a sigmoid layer on top of these networks.

In order to train the models, `binary\_crossentropy' loss function and `RMSprop' optimizer is utilized with learning rate $1e^{-3}$. Training is performed for $50$ epochs with passing $32$ instances in a single iteration. For saving the best intermediate model, we use the Keras callback function.

\subsection{Textual Approach}
Various state-of-the-art transformer models are used including m-BERT \citep{devlin2018bert}, XLNet \citep{yang2019xlnet}, and XLM-Roberta \citep{conneau2019unsupervised} to investigate the textual modality. M-BERT is a multilingual masked language model trained on over $104$ languages. However, during pretraining BERT started with masking a part of the input tokens which eventually creates a fine-tuning discrepancy. This limitation can be alleviated by XLNet model, which is an auto-regressive language model. Using a permutation technique during the training time allows context to consist of both right-side and left-side tokens, making it a generalized order aware language model. In contrast, XLM-Roberta (XLM-R) is introduced for cross-lingual understanding particularly suited for low resource languages \citep{ranasinghe2020multilingual}.
We selected `bert-base-multilingual-cased', `xlnet-base-cased', and `xlm-Roberta-base' models from Pytorch Huggingface\footnote{ https://huggingface.co/transformers/} transformers library and fine-tuned them on our textual data. Implementation is done by using Ktrain \citep{maiya2020ktrain} package. For fine-tuning, we settled the maximum caption length $50$ and used a learning rate of $2e^{-5}$ with a batch size 8 for all models. Ktrain `fit\_onecycle' method is used to train the models for 20 epochs. The early stopping technique is utilized to alleviate the chance of overfitting.     

\subsection{Multimodal Approach}
To verify the singular modality approaches effectiveness, we continue our investigation by incorporating both visual and textual modality into one input data. Various multimodal classification tasks adopted this approach \citep{pranesh2020memesem}. We have employed an early fusion approach \citep{duong2017multimodal} instead of using one modality feature to accept both modalities as inputs and classify them by extracting suitable features from both modalities. For extracting visual features, CNN is used, whereas bidirectional long short term memory (BiLSTM) network is applied for handling the textual features. Different multimodal models have been constructed for the comparison. However, due to the high computational complexity, incorporation of transformer models with CNN-Image models are not experimented in this work.   

\paragraph{CNNImage + BiLSTM:}
Firstly, CNN architecture is employed to extract the image features. It consists of four convolutional layers consisting of $32$, $64$, $128$, and $64$ filters with a size of $3\times3$ in $1^{st}$-$4^{th}$ layers. Each convolutional layer followed by a maxpool layer with a pooling size of $2\times2$. An FC layer with 256 neurons and a sigmoid layer is added after the flatten layer. On the contrary, Word2Vec \citep{mikolov2013distributed} embedding technique applied to extract features from the captions/texts. We use the Keras embedding layer with embedding dimension $100$. A BiLSTM layer with $128$ cells is added at the top of the embedding layer to capture long-term dependencies from the texts. Finally, the output of the BiLSTM layer is passed to a sigmoid layer. After that, the two constructed models' output layers are concatenated together and created a new model. 

\paragraph{Inception + BiLSTM:} 
We combined the pre-trained Inception and BiLSTM network for the classification using both modalities. The inception model is fetched from Keras library and used it as a visual feature extractor. By excluding the top layers, we fine-tuned it on our images with one additional FC layer of 256 neurons and a sigmoid layer. For textual features, similar BiLSTM architecture is employed (Described in CNNImage + BiLSTM). The Keras concatenation layer used two models output layers and combined them to create one model.  
\begin{table*}[h!]
\centering
\begin{tabular}{cl|ccc}
\hline
\textbf{Approach} &\textbf{Classifiers} &\textbf{Precision}&\textbf{Recall}&\textbf{$f_1$-score}\\
\hline   
\multirow{3}{*}{Visual}& CNN & 0.604 & 0.597 & 0.463  \\
& VGG16 &  0.526 & 0.588 & 0.461   \\ 
& Inception & 0.625 & 0.597  & 0.458  \\          
                   
\hline            
\multirow{3}{*}{Textual}& m-BERT &  0.574 & 0.597 & 0.558  \\              
& XLM-R & 0.578  & 0.597  & 0.571   \\ 
& XLNet & 0.592  & 0.609  & \textbf{0.583}   \\   
\hline

\multirow{3}{*}{Multimodal}& CNNImage + BiLSTM & 0.551 & 0.585 & 0.525  \\                    
& ResNet50 + BiLSTM  & 0.669  & 0.603 &  0.471 \\ 
&Inception + BiLSTM & 0.571 & 0.594 &   0.559  \\  

\hline
\end{tabular}
\caption{\label{result} Performance comparison of different models on test set.}
\end{table*}

\paragraph{ResNet50 + BiLSTM:} 
Pretrained residual network (ResNet) \citep{he2015deep} is employed to extract visual features. ResNet  is an extremely deep neural networks with over $150$ layers. Utilizing the skip connection between the layers eradicates the vanishing gradient problem that frequently occurs in a large stacked deep neural network. 
The model is taken from Keras library. To fine-tuned it on our images, the final pooling and FC layer of the ResNet model is excluded. Afterwards, we have added a global average pooling layer with a fully connected layer and a sigmoid layer at the ResNet model's top. A model is constructed for textual features by employing two BiLSTM layers of $128$ and $64$ cells at the top of a word embedding layer with embedding dimension $100$. To mitigate the chance of overfitting a dropout layer with a dropout rate of $0.2$ is introduced between the two BiLSTM layers. The output of the last LSTM layer then transferred to a sigmoid layer. In the end, the output layer of the two models is concatenated to create one combined model.

In all cases, the output prediction is obtained from a final sigmoid layer added just after the concatenation layer of a multimodal model. All the models have compiled using `binary\_crossentropy' loss function. Apart from this, we use `Adam' optimizer with a learning rate of $1e^{-3}$ and choose the batch size of $32$. The models are trained for $50$ epochs utilizing the Keras callbacks function to store the best model. 
\begin{figure*}[h!]
\begin{multicols}{3}
    \subfigure[Inception+BiLSTM]{\includegraphics[height=4cm, width=0.30\textwidth]{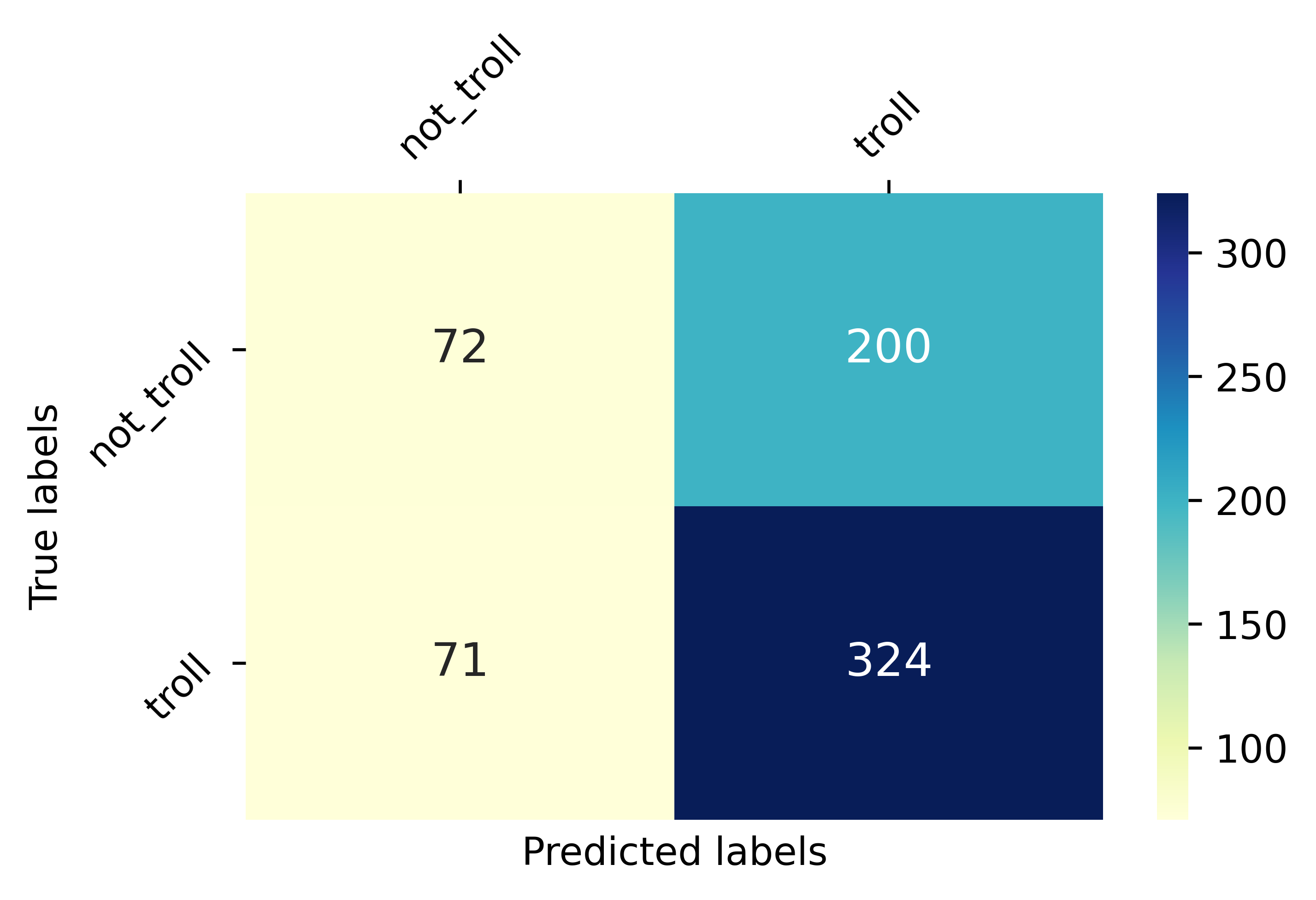}}
    \subfigure[Inception]{\includegraphics[height=4cm, width=0.30\textwidth]{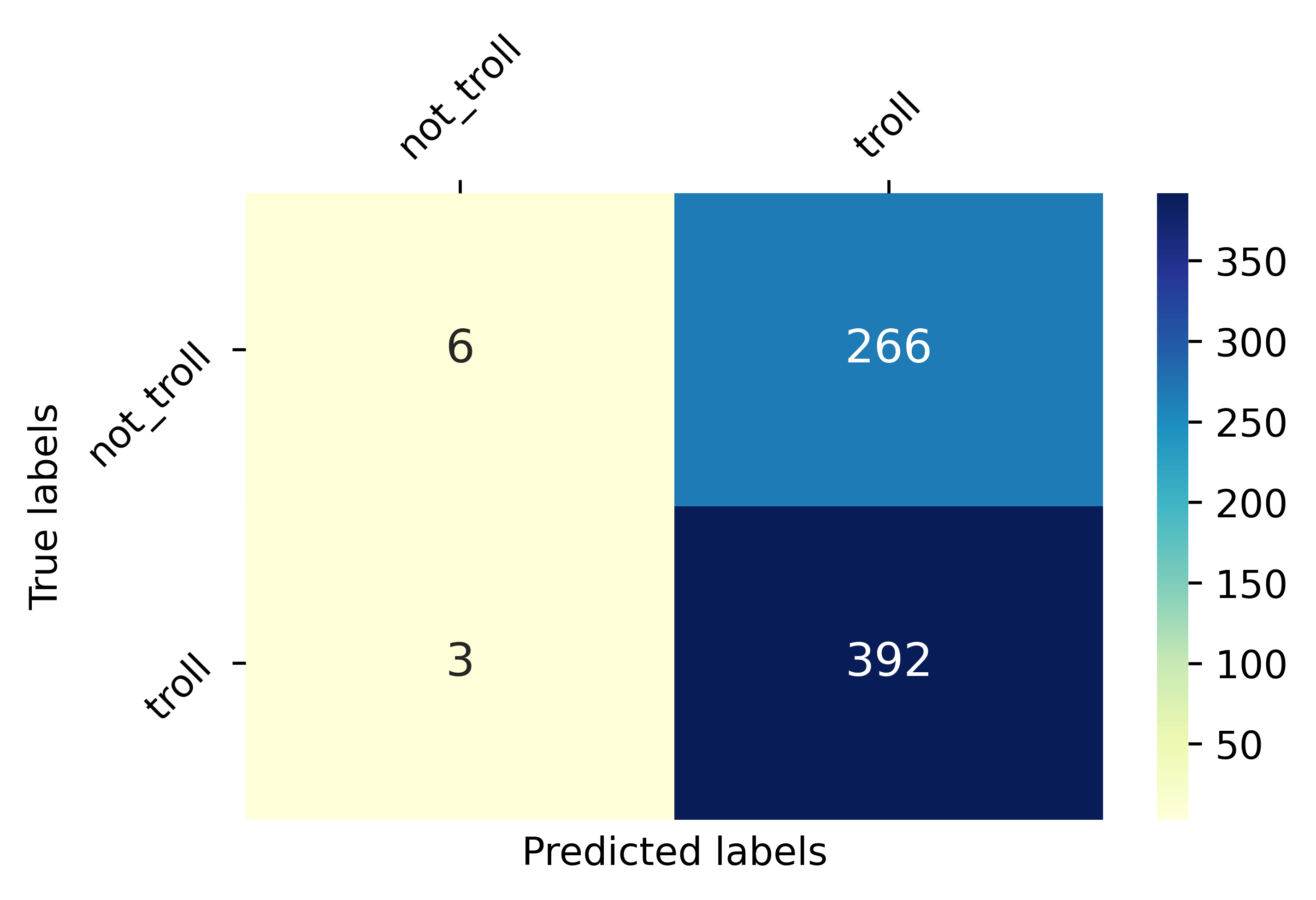}}
    \subfigure[XLNet]{\includegraphics[height=4cm, width=0.30\textwidth]{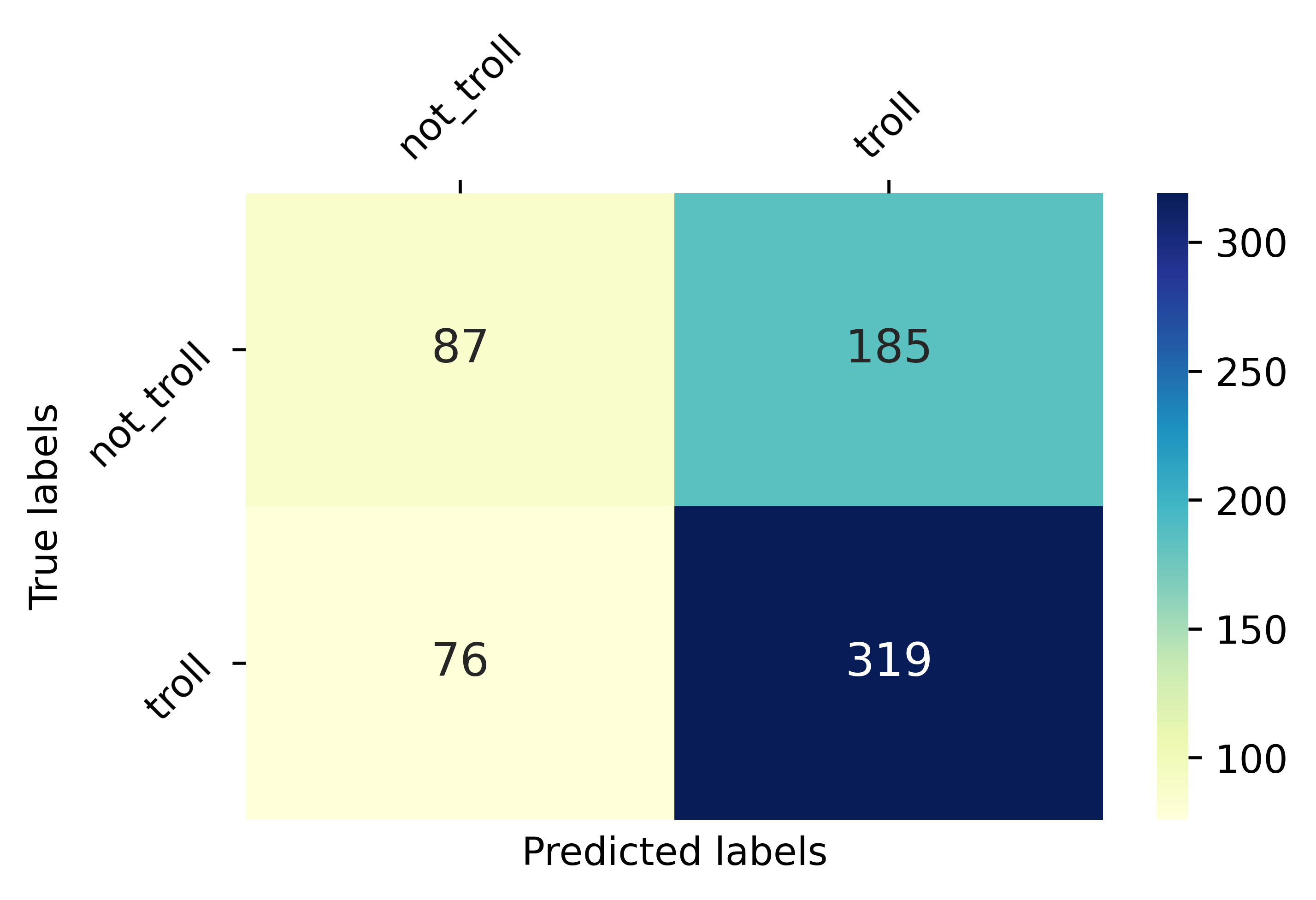}}
   
    \end{multicols}
 \caption{Confusion Matrix of the best model in each approach (a) Multimodal  (b) Visual (c) Textual}
 \label{confusion}
\end{figure*}

\section {Result and Analysis}
\label{section5}
This section presents a comparative performance analysis of the different experimental approaches to classify the multimodal memes. The superiority of the models is determined based on the weighted $f_1$-score. However, other evaluation metrics like precision and recall are also considered in some cases. 
Table~\ref{result} shows the evaluation results of different approaches on the test set. The outcome reveals that all the models (i.e. CNN, VGG16, and Inception) developed based on the imaging modality obtained approximately same $f_1$-score of $0.46$. However, considering both precision ($0.625$) and recall ($0.597$) score, only Inception model performed better than the other visual models. On the other hand, models that are constructed based on textual modality show a rise of $10-12\%$ in $f_1$ score than the visual approach. In the textual approach, m-BERT and XLM-R achieved $f_1$-score of $0.558$ and $0.571$ respectively. However, XLNet outperformed all the models by obtaining the highest $f_1$-score of $0.583$. For comparison, we also perform experiment by combining both modality features into one model. In case of multimodal approach, CNNImage + BiLSTM model obtained $f_1$-score of $0.525$ while ResNet + BiLSTM model achieved a lower $f_1$- score ($0.47$) with a drops of $6\%$. Compared to that Inception + BiLSTM model achieved the highest $f_1$-score of $0.559$. Though two multimodal models (CNNImage + BiLSTM and Inception + BiLSTM) showed better outcome than visual models, they could not beat the textual model’s performance (XLNet). Though it is skeptical that XLNet (monolingual model) outperformed multilingual models (m-BERT and XLM-R), however the possible reason might be due to the provided captions which were actually written in English.

From the above analysis, it is evident that visual models performed poorly compared to the other approaches. The possible reason behind this might be due to the overlapping of multiple images in all the classes. That means the dataset consists of many memes which have same visual content with different caption in both classes. Moreover, many images do not provide any explicit meaningful information to conclude whether it is a troll or not-troll meme.

\subsection{Error Analysis}
A detail error analysis is performed on the best model of each approach to obtain more insights. The analysis is carried out by using confusion matrix (Figure~\ref{confusion}). From the figure~\ref{confusion} (a), it is observed that among $395$ troll memes, Inception + BiLSTM model correctly classified $324$ and misclassified $71$ as not-troll. However, this model’s true positive rate is comparatively low than the true negative rate as it identified only $72$ not-troll memes correctly and wrongly classified $200$ memes. On the other hand, in the visual approach, the Inception model showed outstanding performance by detecting $392$ troll memes correctly from $395$. However, the model confused in identifying not-troll memes as it correctly classified only $6$ memes and misclassified $266$ from a total $272$ not-troll memes. Meanwhile, figure~\ref{confusion} (c) indicates that among $272$ not-troll memes, XLNet correctly classified only $87$. In contrast, among $395$, the model correctly identified $319$ troll memes.   

The above analysis shows that all models get biased towards the troll memes and wrongly classified more than $70\%$ memes as the troll. The probable reason behind this might be the overlapping nature of the memes in all classes. Besides, many memes do not have any embedded captions which might create difficulty for the textual and multimodal models to determine the appropriate class. Moreover, we observed that most of the missing caption memes were from troll class which might be strong reason for the text models to incline towards the troll class.

\section{Conclusion}
This work present the details of the methods and performance analysis of the models that we developed to participate in the meme classification shared task at EACL-2021. We have utilized visual, textual and multimodal features to classify memes into the `troll’ and `not troll’ categories. Results reveal that all the visual classifiers achieve similar weighted $f_1$-score of $0.46$. Transformer-based methods capture textual features where XLNet outdoes all others by obtaining $0.58$ $f_1$-score. In the multimodal approach, visual and textual features are combined via early fusion of weights. $F_1$ score rose significantly after adding textual features in CNN and Inception models. Only BiLSTM method is applied to extract features from the text in this approach. In future, it will be interesting to see how the models behave if transformers use in place of BiLSTM. An ensemble of transformers might also be explored in case of textual approach.

\bibliography{anthology,eacl2021}
\bibliographystyle{acl_natbib}

\end{document}